\def\year{2019}\relax
\documentclass[letterpaper]{article} 
\usepackage{aaai19}  
\usepackage{times}  
\usepackage{helvet}  
\usepackage{courier}  
\usepackage{url}  
\usepackage{graphicx}  

\usepackage{multirow}
\usepackage{array}
\usepackage[pdfencoding=auto]{hyperref}
\usepackage{amsmath,amssymb}
\usepackage{diagbox}
\usepackage{arydshln}
\usepackage{rotating}
\usepackage{caption}
\usepackage{comment}
\usepackage{hhline}

\newcommand*\samethanks[1][\value{footnote}]{\footnotemark[#1]}

\begin{document}
%
\title{Generating Diverse and Accurate Visual Captions by Comparative Adversarial Learning}

\author{Dianqi Li\textsuperscript{\rm 1}\thanks{The work was done while Dianqi Li and Xiaodong He were at Microsoft Research.}, 
Qiuyuan Huang\textsuperscript{\rm 2}, 
Xiaodong He\textsuperscript{\rm 3}\samethanks, 
Lei Zhang\textsuperscript{\rm 2}, 
Ming-Ting Sun\textsuperscript{\rm 1}\\
\textsuperscript{\rm 1}University of Washington, \textsuperscript{\rm 2}Microsoft Research, \textsuperscript{\rm 3}JD AI Research\\ 
\{dianqili, mts\}@uw.edu, xiaodong.he@jd.com, \{leizhang, qihua\}@microsoft.com\\
}

\maketitle
\begin{abstract}
We study how to generate captions that are not only accurate in describing an image but also diverse across different images. The problem is both fundamental and interesting, as most machine-generated captions, despite phenomenal research progresses in the past several years, are expressed in a very monotonic and featureless format. While such captions are normally accurate, they often lack important characteristics in human languages - distinctiveness for each image and diversity across different images. To address this problem, we propose a novel conditional generative adversarial network for generating diverse captions across images. Instead of estimating the quality of a caption solely on one image, the proposed comparative adversarial learning framework better assesses the quality of captions by comparing a set of captions within the image-caption joint space. By contrasting with human-written captions and image-mismatched captions, the caption generator effectively exploits the inherent characteristics of human languages, and generates more diverse captions. We show that our proposed network is capable of producing accurate and diverse captions across images.
\end{abstract}

\section{Introduction}

\begin{figure}[t]
\centering
\includegraphics[width=1.0\linewidth]{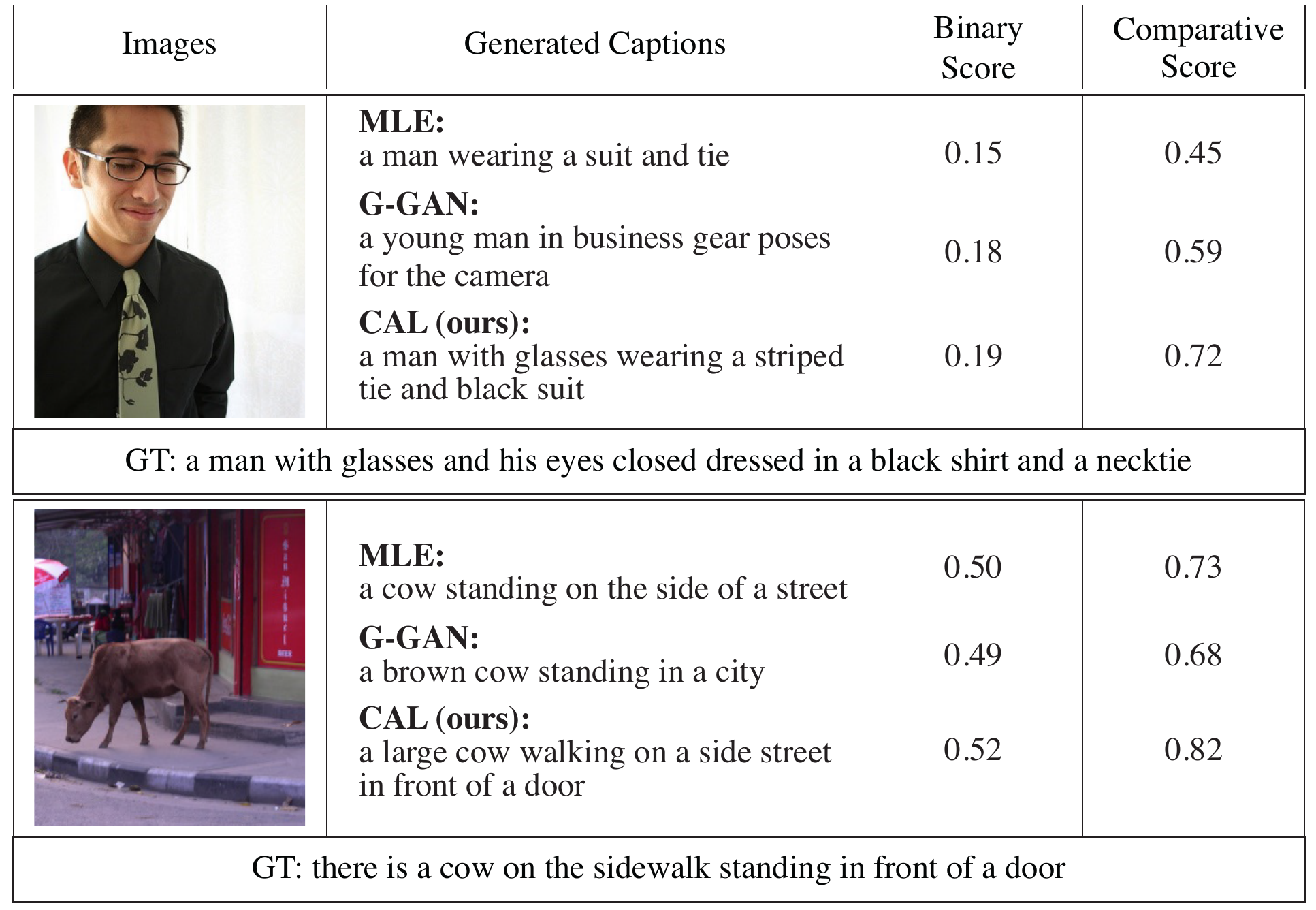}
\caption{Captions generated by MLE, conditional GANs (G-GAN) with binomial scores and our comparative adversarial learning network (CAL) with comparative scores. The shown scores are evaluated by the discriminators in G-GAN and CAL, respectively. The proposed adversarial framework estimates comparative scores by comparing a collection of captions,  leading to more accurate and discriminative rewards for caption generator.}
\label{fig:fig1}
\end{figure}

Image caption generation has attracted great attentions due to its wide applications in many fields, such as semantic image search, image commenting in social chat bot, and assistance to visually impaired people.  Benefiting from recent advancements of deep learning, most existing works employ convolutional neural networks (CNNs) and deep recurrent language models, and have achieved great performance improvement on automatic evaluation metrics, such as BLEU~\cite{papineni2002bleu}, CIDEr~\cite{vedantam2015cider}, etc. 

Despite such successes, machine-generated captions are often in a generic format and can be easily differentiated from human-written captions, which tend to be more descriptive and diverse. As most state-of-the-art image caption algorithms are learning-based, to best match with the ground truth captions, such algorithms often produce high-frequency n-gram patterns or common expressions. As a result, the generated image captions receive high scores on automatic evaluation metrics, yet lack a significant characteristic in human language - diversity across different images. From human perspectives, as demonstrated in~\cite{jas2015image}, each image possesses its own specificity, and accordingly its related captions should acquire its distinctiveness, leading to diverse captions for different images. In general, distinctive descriptions are often pursued by human, who can easily distinguish a specific image among a group of similar images.
In this work, our goal is to generate diverse and accurate captions which are similar to human-written descriptions.

Recent success of Generative Adversarial Networks (GANs)~\cite{mirza2014conditional} provides a possible way to generate diverse captions~\cite{Dai2017Towards,shetty2017speaking}. In this setting, a caption generator and a discriminator are jointly trained by a binomial distribution, which estimates the relevance and quality of the captions to the image. However, due to the large variability of natural language, a binary predictor is usually incapable of representing the richness and diversity of captions. To ensure semantic relevance, a regularization term for distinguishing mismatched captions must be included during training. 

In contrast to assigning an absolute score to a caption for one image, we noticed that it is relatively easier to distinguish the qualities of two captions by comparison. Motivated by this, we propose a comparative adversarial learning (CAL) network to learn human-like captions. Specifically, contrary to an absolute binary score for one caption, the quality of the caption is assessed relatively by comparing it with other captions in the image-caption space. In adversarial learning, the proposed discriminator ranks the human references, which are more specific and distinctive, higher than generic captions that have high-frequency n-gram patterns or common expressions. Consequently, with the guides from the discriminator, the generator effectively learns to generate more specific and distinctive captions, hence increases the diversity across the corpus. In summary, our main contributions lie in three aspects:
\begin{itemize}
\item We propose a novel comparative adversarial learning network, which is capable of generating more diverse and better captions across images by comparing different captions.
\vspace{-1mm}
\item By suppressing the scores of image-mismatched captions, especially for those from similar images, the proposed comparative learning framework can inherently ensure semantic relevance without involving an regularization term for mismatched captions.
\vspace{-1mm}
\item To effectively measure the caption diversity across images, we propose a new metric based on the semantic variance from caption embedding features. Additionally, experimental results clearly demonstrate the effectiveness of the proposed framework in terms of diversity and quality.
\vspace{-1mm}
\end{itemize}
\section{Related Work }
\textbf{Diverse Image Captioning}\hspace{2mm} 

\noindent Most image captioning systems use an encoder-decoder framework which shares a similar idea as sequence learning~\cite{sutskever2014sequence,gehring2017convolutional,vaswani2017attention,peng2018rational,peng2019text}. Typically, the networks are trained by maximum likelihood estimation (MLE)~\cite{vinyals2015show,karpathy2015deep,xu2015show,SCN_CVPR2017} or reinforcement learning~\cite{Rennie2016Self,ren2017deep,liu2017improved,Anderson2017up-down,luo2018discriminability,liu2018show}. Although such methods achieve outstanding performances on conventional evaluation metrics, such as BLEU, CIDEr, etc., the generated captions usually consist of high-frequency n-gram patterns but lose the diversity across images and thus are unnatural to human. 
To remedy this weakness, diverse beam search and ensemble methods~\cite{vijayakumar2016diverse,wang2016diverse} have been proposed.~\cite{wang2017diverse,chatterjee2018diverse} work on diverse image captioning by using variational auto-encoders. To achieve better caption diversity,~\cite{Dai2017Towards,shetty2017speaking} incorporate generative adversarial nets (GANs) into image captioning systems, with a binary-based discriminator. However, in sequence adversarial training, a binary-based discriminator is easily trained much stronger than the generator~\cite{chemaximum,guo2017long,lin2017adversarial}, resulting in less distinguishable rewards or gradient vanishing problems for the generator (\hyperref[fig:fig1]{Figure 1}). To generate captions with correct semantic relevance,~\cite{Dai2017Towards,shetty2017speaking} must train the binary discriminator under an additional regularization.

In this work, we propose a comparative adversarial learning framework that explicitly estimates the quality of captions in a more discriminative way, which in turn helps the generator to produce more diverse captions while maintaining the caption correctness without the additional discriminator regularization.
\vspace{2mm}

\begin{figure}[t]
\centering
\includegraphics[width=1.0\linewidth]{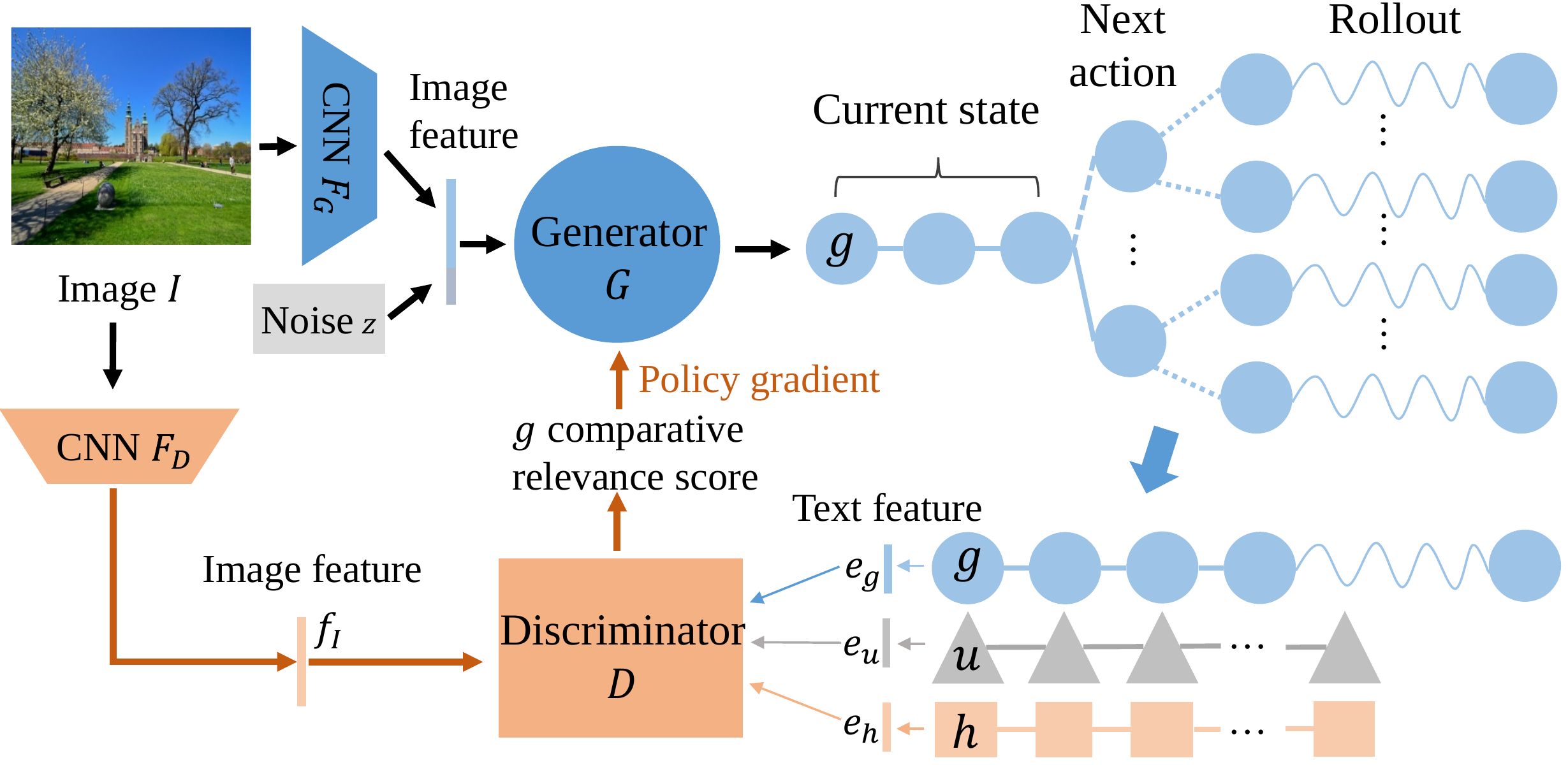}
\caption{Comparative Adversarial Learning Network. The discriminator $D$ is trained over comparative relevance scores for each image by comparing a generated caption $g$, a human-written caption $h$, and unrelated captions $u$. The generator $G$ is optimized by policy gradient where the reward is  estimated by the expectation of the comparative relevance score $g$ over K rollout simulation captions.}
\label{fig:fig2}
\end{figure}

\noindent\textbf{Diversity Metrics}\hspace{2mm} 

\noindent Automatic evaluation metrics such as BLEU, CIDEr-D, etc., have been widely applied for evaluating the quality of generated captions. Nonetheless, the evaluation of diversity across captions is still an open problem. Human language, inherited immense complexity and sophisticated interpretation, poses a thorny problem for developing standard criterion. 
~\cite{li2015diversity,vijayakumar2016diverse,jain2017creativity} measure the degree of diversity by analyzing distinct n-grams or word usages for generated sentences with respect to ground truths. This reflects an inventiveness for generated sentences, but not a diversity aspect among all the generated sentences. To estimate the caption diversity at the token level,~\cite{shetty2017speaking,wang2017diverse,deshpande2018diverse} inspect n-gram usage statistics and the size of vocabulary in all generate captions. However, the diversity of sentences is not only represented by various word or phrase usages, but also variant long-term sentence patterns and even implications of sentences. A simple n-gram statistics is unable to assess the diversity at the sentence level. In this paper, we propose a novel diversity metric based on semantic sentence features which compensate the defects of previous methods.
\section{Comparative Adversarial Learning Network}
As shown in \hyperref[fig:fig2]{Figure 2}, the proposed Comparative Adversarial Learning (\textbf{CAL}) Network consists of a caption generator $G$ and a comparative relevance discriminator (\textbf{cr-discriminator}) $D$. The two subnetworks play a min-max game as follow:
\begin{align}
\min_{\theta} \max_{\phi} \mathcal{L}(G_{\theta}, D_{\phi}), 
\end{align}
in which $\mathcal{L}$ is an overall loss function, while $\theta$ and $\phi$ are trainable parameters in $G$ and $D$, respectively. Given a reference image $I$, the generator $G_\theta$ outputs a sentence $g$ as the corresponding caption. $D_\phi$ aims at correctly estimating the comparative relevance score (\textbf{cr-score}) of $g$ with respect to human-written caption $h$ within the image-caption joint space. $G_\theta$ is trained to maximize the cr-score of $g$ and generate human-like descriptions trying to confuse $D_\phi$. We will elaborate each subnetwork in the following sections.

\subsection{Caption Generator}
Our caption generator $G_\theta$ is based on the standard encoder-decoder architecture~\cite{vinyals2015show}. The captioning image encoder model $F_G$ first extracts a fixed dimensional feature from image $I$ using a CNN. Then a text decoder, implemented by a long short-term memory (LSTM) network, interprets the encoded feature $F_G(I)$ into a word sequence $g_{0:T} = (g_0, ..., g_{T})$ to describe image $I$, where $g_t$ is a token in time step $t$ and $T$ is the maximum time step. To produce captions with more variations, the encoded feature $F_G(I)$ can be concatenated with a random vector $z$. The notation of $z$ will be ignored in the rest parts for simplicity. In time step $t$ generation, the next token $g_t$ can be sampled by:
\begin{align}
g_t \sim \pi_\theta(g_t | I, g_{0:t-1}), \hspace{3mm} t \in (1, T)
\end{align}
$\pi_\theta$ is a word distribution, determined by inputs and $\theta$, over all the words in vocabulary $V$. By sequentially sampling or greedy decoding words according to $\pi_\theta$, a complete caption $g_{1:T}$ can be generated by captioner $G_\theta$. In comparative adversarial training, $G_\theta$  expects to produce better captions with higher cr-scores. However, unlike cross-entropy loss in the MLE method, the cr-score of $g_{1:T}$ estimated by discriminator $D_\phi$ is based on discrete tokens, whose gradients cannot be directly employed for $G_\theta$ through back-propagation. Therefore, we adopt a common technique - Policy Gradient method~\cite{sutton2000policy} to solve this gradient issue. The details will be discussed in Section 3.3. 

\subsection{Comparative Relevance Discriminator}
\cite{Dai2017Towards} propose to estimate the semantic relevance, naturalness, and quality of a generated caption by a logistic function over the similarity between the caption and the given image. However, an absolute binary value is very restrictive to evaluate them all, especially the quality of a caption. To evaluate a discriminative score, it is more justifiable to compare a generated caption with other captions, primarily with human-written caption $h$. Therefore, we formulate a comparative relevance score (\textbf{cr-score}) to measure an overall image-text quality of caption $c$ by comparing a set of captions $\mathcal{C}^c$ given image $I$:
\begin{align}
D_{\phi}(c | I, \mathcal{C}^c) = \frac{exp(\gamma S(e_c, f_I)}{\sum_{c^{'} \in \mathcal{C}^c}exp(\gamma S(e_{c^{'}}, f_I))}
\label{eq3}
\end{align}
where $\mathcal{C}^c$ denotes a set of captions including $c$, and the cr-score of $c$ is what we care about here. $e_c$ and $f_I$ are the text feature and image feature extracted by the text encoder and CNN image encoder $F_{D_\phi}$ in discriminator $D_\phi$, respectively. The cosine similarity between $e_c$ and $f_I$ is defined as $S(e_c, f_I) = (e_c^{T}f_I) / (\lVert e_c \rVert \lVert f_I \rVert)$. $\gamma$ is an empirical parameter defined by validation experiment. A higher $\gamma$ leads $D_{\phi}(c | I, \mathcal{C}^c)$ towards the caption that better matches with image $I$. 
\begin{figure}[t]
\centering
\includegraphics[width=1.0\linewidth]{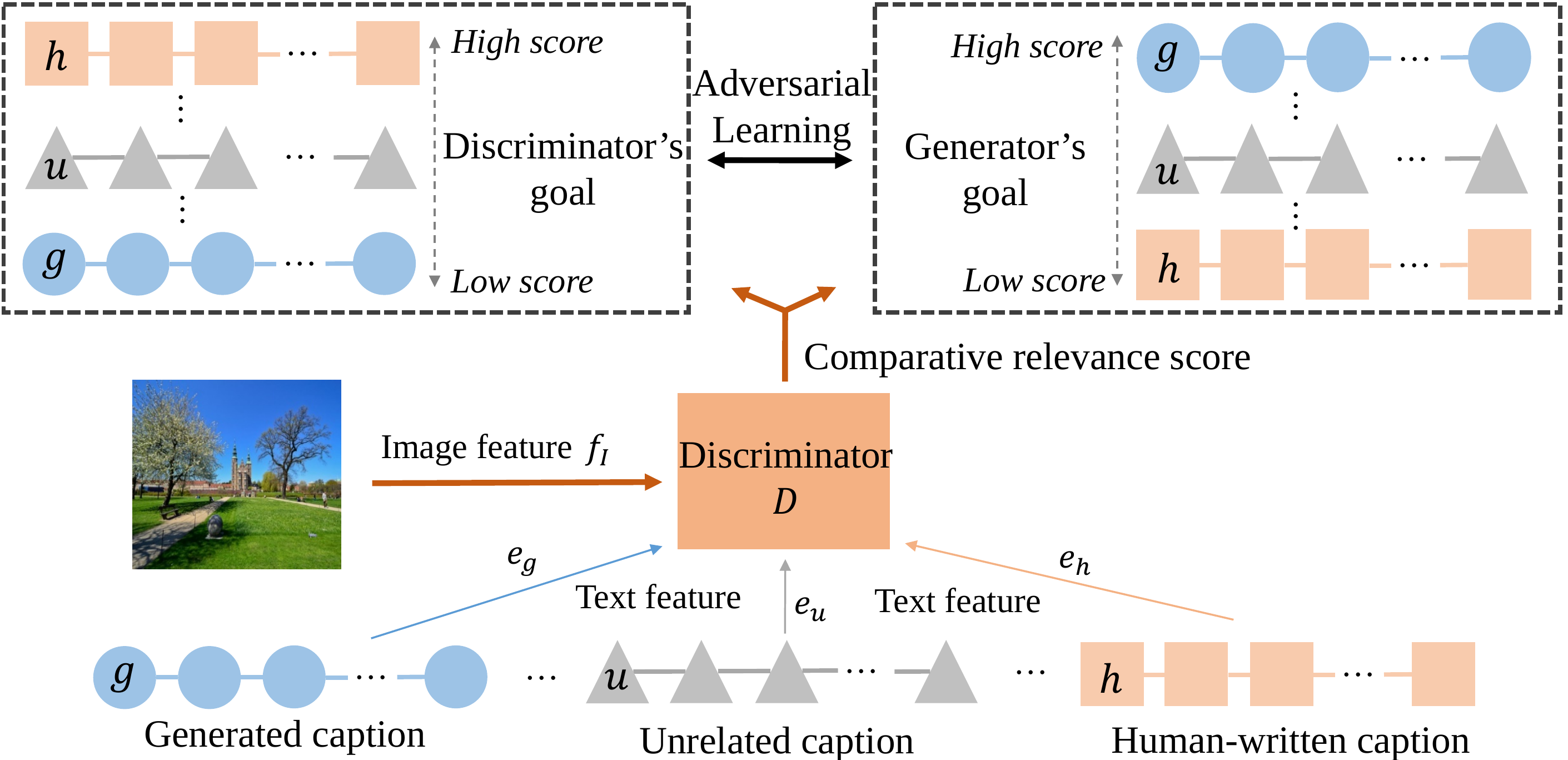}
\caption{Training objectives in our adversarial learning. While the discriminator desires to judge human-written captions correctly with higher cr-scores, the generator aims to produce captions with higher cr-scores and thus confusing the discriminator.} 
\label{fig:fig3}
\end{figure}
$D_{\phi}(c | I, \mathcal{C}^c))$ estimates the cr-score of caption $c$ by comparing with other captions in the image-caption joint space - a higher score represents caption $c$ is superior in $\mathcal{C}^c$. To obtain more accurate cr-score for $c$, it is favorable to include human-written caption $h$ for image $I$ in $\mathcal{C}^c$. In this case, the cr-score of $c$ contains a discrepancy information between caption $c$ and human-written caption $h$. The discriminator is designed to differentiate generated captions from human-written captions for image $I$. Specifically, from the discriminator's perspective, a human-written caption  desires to receive a higher cr-score, whereas a generated caption should receive a lower cr-score (\hyperref[fig:fig3]{Figure 3}). Hence, the objective function to be maximized for discriminator can be defined as:
\begin{equation}
\mathop{\mathbb{E}}_{h \sim \mathcal{P}_h}\left[ \log D_{\phi}(h | I, \mathcal{C}^h) \right] +
\mathop{\mathbb{E}}_{g \sim G_\theta} \left[\log(1- D_{\phi}(g | I, \mathcal{C}^g) \right]
\label{eq_obj_discriminator}
\end{equation}
where $\mathcal{P}_h(I)$ represents human-written caption distribution given image $I$. Set $\mathcal{C}^h$ and $\mathcal{C}^g$ encloses a human-written caption $h$, a machine-generated caption $g$, and other unrelated captions $u$. In experiments, $u$ can be directly obtained from image-mismatched captions in one mini-batch. 

\subsection{Policy Gradient Optimization for \texorpdfstring{$G_\theta$}{Lg}}
In contrast to $D_\phi$, the caption generator $G_\theta$ attempts to maximize the cr-scores of machine-generated captions and thus fool the discriminator (\hyperref[fig:fig3]{Figure 3}). However, the cr-scores of a generated caption $g$ are assessed by $D_\phi$ based on a series of sequential discrete samples, which are non-differentiable during training.  We address this problem by a classic policy gradient method~\cite{sutton2000policy}. Considering in each time step $t$, the generation of each word $g_t$ is an action of an "agent" $G_\theta$ from policy $\pi_\theta$ according to the current state $(I, g_{0 : t-1})$.  An intermediate reward $r$ for this action is approximated as the expected future reward:
\begin{align}
Q_\theta (g_t | I, g_{0 : t-1}) = \mathop{\mathbb{E}}_{g_{t+1:T}}[r(g_{0 : t-1}, g_t, g_{t+1:T} | I)]
\label{eq:4}
\end{align}
The action reward $r$ can be any metric, including the cr-score from $D_\phi$. Unfortunately, the discriminator cannot provide a score unless a complete sentence is generated. The lack of intermediate rewards will result in a gradient vanishing problem. To imitate an accurate intermediate reward, following~\cite{yu2017seqgan}, we deploy a $K$-times Monte Carlo rollout process conditioned on the current caption generator $G_\theta$ to explore the rest unknown words $g_{t+1: T}$. Then the intermediate reward for action $g_t$ can be approximated by the expected cr-score over $K$ rollout simulation captions:
\begin{align}
\begin{array}{c}
Q_{\theta, \phi} (g_t | I, g_{0 : t-1}) \simeq \frac{1}{K} \sum\limits_{k = 1}^{K} D_\phi(g_{k, t} | I, C^{g_{k, t}})
\end{array}
\end{align}
where $ g_{k, t} = g_{0:t-1}, g_t \oplus \bar{g}_{t+1, T}$, and $\bar{g}_{t+1, T} $ is sampled from $G_\theta$ by the rollout method. Besides, $t \in (1, T-1)$, as $g_0$ is a start token and $G_\theta$ receives an accurate reward once generating a full sequence. $C^{g_{k, t}}$ contains a simulated caption $g_{k, t}$, a human-written caption $h$ and other unmatched descriptions $u$, corresponding to the image $I$. To train the generator, the objective is to optimize the policy and adjust the generator to receive a maximum long-term reward - higher cr-scores for generated captions in each time step (Fig.~\ref{fig:fig3}). In the end, the gradient for updating generator $G_\theta$ can be finalized by:
\begin{align}
\mathop{\mathbb{E}}_{g \sim G_\theta}   \sum_{t = 1}^{T}  \nabla_\theta \pi_\theta(g_t | I, g_{0:t-1}) \cdot  Q_{\theta, \phi} (g_t | I, g_{0 : t-1})
\end{align}
\noindent where $g_t$ is an intermediate token belonging to $g$ at time step $t$. The goal of the generator is to maximize the expected cr-scores of generated captions. 

\subsection{Comparisons with Previous Models}
During discriminator training, ~\cite{Dai2017Towards,shetty2017speaking} introduce a regularization term to learn image-caption matching by minimizing binary scores of mismatched captions $u$ (last term in the below equation):
\begin{equation}
\small{
\mathop{\mathbb{E}}_{h,g,u} \log D_{b}(h | I) +
\log (1 - D_{b}(g | I)) +
\log (1 - D_{b}(u | I))
}
\label{eq8}
\end{equation}
\normalsize
\noindent where $D_b$ is a binary discriminator. However, the cr-discriminator $D_\phi$ can naturally learn such image-caption matching by placing mismatched captions in the comparison set $C^h$ with true captions $h$ and generated captions $g$. Specifically, by enlarging the cr-score of the matched image-caption pair $(h, I)$ in set $C^h$, $D_\phi$ can consistently distinguish its corresponding caption from others, and suppress the scores for mismatched descriptions $u$ (\hyperref[eq3]{Equation 3}). $D_\phi$ can in turn help the caption generator $G_\theta$ produce diverse captions for corresponding images, ensuring semantic relevances of generated captions. Meanwhile, the binary discriminator $D_b$ separates the decisions on $g$ and $h$. The proposed network simply combines the two separate decisions into a single ranking process. The cr-score of the generated captions are estimated by contrasting human-written sentences subject to image $I$. This can assist the cr-score to comprise more informative guidances, including both naturalness and quality from ground truths, benefiting the training of the caption generator $G_\theta$. 
\section{Experiments}

\begin{table}[t]
\small
\begin{center}
\setlength{\tabcolsep}{5.1pt}
\begin{tabular}{cccccc}\hline
      \noalign{\smallskip}
      Model & {\scriptsize BLEU4} & {\scriptsize METEOR} & {\scriptsize ROUGE} & {\scriptsize CIDEr} & {\scriptsize SPICE} \\ \hline
      \noalign{\smallskip}
        Human &0.190 &0.240 &0.465 &0.861 &\textbf{0.208} \\
      \noalign{\smallskip}
        MLE &\textbf{0.297} &\textbf{0.252} &\textbf{0.519} &\textbf{0.921} &0.175 \\
      \noalign{\smallskip}
        G-GAN &0.208 &0.224 &0.467 &0.705 &0.156 \\
      \noalign{\smallskip}
        CAL (ours)  &0.213 &0.225 &0.472 &0.721 &0.161 \\ \hline
        \vspace{0.01mm}
      \end{tabular}
\caption{Performance comparisons on MSCOCO test set. In human result, a sentence randomly sampled from ground-truth annotations is evaluated by the rest annotations for each image.}
\label{table:tab1}
\end{center}
\end{table}

\noindent\textbf{Models.}\hspace{2mm}To test the effectiveness of the proposed Comparative Adversarial Learning (CAL) network, we compare two baseline models:

\noindent
(1) \textit{MLE}: We use LSTM-R~\cite{SCN_CVPR2017} based on the mainstream CNN-LSTM architecture as our MLE baseline model. The training follows the standard MLE method.

\noindent
(2) \textit{Adversarial models}: We use \textit{G-GAN}~\cite{Dai2017Towards} as the baseline model for diverse image captioning (\textbf{G} represents the generator). The corresponding discriminator $D_b$ outputs a binary score in $[0, 1]$ through a logistic function over the dot product between image and text features (\hyperref[eq8]{Equation 8}).

To make a fair comparison, all image features for generators and discriminators are extracted by \textit{ResNet-152}~\cite{he2016deep} (we reimplement \textit{G-GAN} by using \textit{ResNet-152} network as image encoders). All text-decoders in generators and text-encoders in discriminators are implemented by LSTMs. More details related to the networks are included in~\hyperref[sec:appendix1]{Appendices A.1}

\vspace{2mm}
\noindent\textbf{Training.}\hspace{2mm} Before adversarial training, the caption generator $G_\theta$ in both adversarial models is pretrained by the standard MLE method~\cite{vinyals2015show}~\cite{karpathy2015deep} for 20 epochs, and the cr-discriminator is pretrained according to \hyperref[eq_obj_discriminator]{Equation 4} for 10 epochs. During the experiment, we found the generator pretraining is necessary, otherwise it will encounter mode collapse problem and generate nonsense captions. On the other hand, pretraining discriminator helps more stable training later. In the adversarial learning stage, two sub-networks are trained jointly, in which every one generator iteration is followed by 5 discriminator iterations. We set the learning rate to $0.0005$ and the batch size to $64$. In each mini-batch with 64 image-caption pairs, all other 63 captions that are not corresponding to the correct image are used as the unrelated captions during training. The rollout number $K$ is empirically set to 16, and $\gamma$ is set to $10$. During testing, the generated captions are sampled based on policy and the one with the best cr-score is chosen for evaluation.

\vspace{2mm}
\noindent\textbf{Dataset.}\hspace{2mm} We conduct all experiments on the MSCOCO dataset~\cite{lin2014microsoft}. MSCOCO contains 123,287 images, each being annotated with at least 5 human-written captions. All our experiments are based on the public split method from~\cite{karpathy2015deep}: 5000 images for both validation and testing, and the rest for training. 

\begin{figure}[t]
\centering
\includegraphics[width=1.0\linewidth]{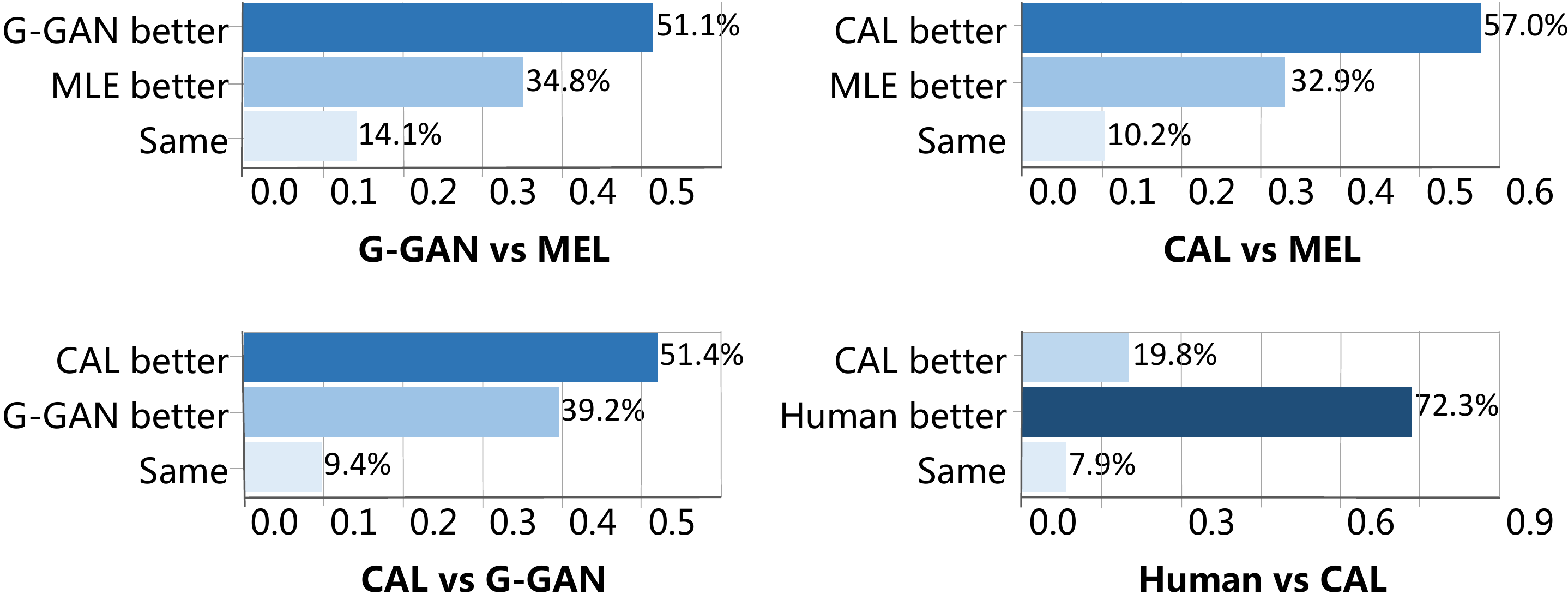}
\caption{Human evaluation results by comparing model pairs. The majority of respondents agree that our proposed CAL generates better captions than the two baselines. The numbers in the figure represent the ratio of total survey cases.}
\label{fig:fig4}
\end{figure}

\vspace{2mm}
\noindent\textbf{Evaluation.}\hspace{2mm} We evaluate the generated captions based on both the correctness and diversity metrics, which guarantee the generation quality in each aspect. While the correctness of the generated captions is measured by common captioning metrics, the diversity across various images is evaluated by the proposed metric based on caption embedding features.

Consider each image is annotated by one caption, whose embedding feature is extracted by a same text encoder. Ideally, all embedding features are identical and the feature variance is zero if all the images have same captions. Conversely, a large variance would present if all the captions were distinct. Thus, the variance across embedding features reflects the diversity of captions on a semantic-level. To measure the variance, all the text embedding features can be concatenated into a feature matrix $A\in \mathbb{R}^{m \times n}$, where $m$ is the number of captions and $n$ is the dimensions of the embedding feature.
To estimate the correlation $\sigma_i$ in each dimension, the covariance matrix $M\in \mathbb{R}^{n \times n}$ of $A$ can be computed. Then, $\sigma_i$ can be obtained by singular value decomposition (SVD): $M = U \Sigma V^{T}$, where $\Sigma = diag(\sigma_0, ..., \sigma_{n-1})$; $U$ and $V^{T}$ are $m \times m$ and $n \times n$ unitary matrix. 
 
Finally, we use $l_1$-norm $\hat{\sigma} = \sum_{i = 0}^{n-1} |\sigma_i|$ to evaluate an overall variance in all dimensions among caption embedding features. A large variance $\hat{\sigma}$  suggests the embedding features of captions are less akin or correlated, representing more distinctive expressions and larger diversity among image captions.

\begin{table}[t]
\small
\begin{center}
\setlength{\tabcolsep}{5.5pt}
\begin{tabular}{ccccc}
\hline\noalign{\smallskip}
Category & MLE & G-GAN & CAL (ours) & Human \\
\noalign{\smallskip}
\hline

\noalign{\smallskip}
Bathroom & 2.733 & 6.145 & 6.501 & \textbf{9.066} \\
\noalign{\smallskip}
Computer & 3.710 & 6.012 & 7.228 & \textbf{8.943} \\
\noalign{\smallskip}
Pizza & 3.837 & 5.779 & 6.805 & \textbf{9.117} \\
\noalign{\smallskip}
Building & 4.019 & 5.940 & 6.088 & \textbf{9.344} \\
\noalign{\smallskip}
Cat & 4.196 & 5.225 & 6.473 & \textbf{9.155} \\
\noalign{\smallskip}
Car &4.968 &5.910 &6.661 & \textbf{8.741} \\
\noalign{\smallskip}
Daily supply & 5.056 & 6.204 & 7.330 & \textbf{9.075} \\
\noalign{\smallskip}
$All^*$ &6.947 &7.759 &8.812 & \textbf{9.465} \\
\noalign{\smallskip}
\hline
\end{tabular}
\end{center}
\caption{ Diversity evaluations across various image categories. $All^{*}$ denotes all the images in MSCOCO test set. }
\label{table:tab2}
\end{table}

\vspace{2mm}
\noindent\textbf{Accuracy.} We first evaluate the generated captions from different models on five automatic metrics: BLEU4~\cite{papineni2002bleu}, METEOR~\cite{banerjee2005meteor}, ROUGE\_L~\cite{lin2004rouge}, CIDEr-D~\cite{vedantam2015cider} and SPICE~\cite{anderson2016spice}. As can be seen in \hyperref[table:tab1]{Table 1}, although our method CAL slightly outperforms the baseline G-GAN, the standard MLE model yields remarkably better results, even outperforms human. However, as discussed by~\cite{Dai2017Towards,shetty2017speaking}, these evaluation metrics overly focus on n-grams matching with ground truth captions and ignore other important human language factors such as diversity. The captions, written with variant expressions, have fewer n-grams matched with ground truths. As a result, captions with novel expressions from the human and adversarial models receive lower scores on these metrics. These metrics particularly represent the quality of pattern matching, instead of the overall quality from human perspective.

\vspace{2mm}
\noindent\textbf{Human Evaluation.}\hspace{1mm} To correlate with human judgments on the correctness of captions, we conducted human evaluation experiments on Amazon Mechanical Turk. Specifically, we randomly sampled 300 images from test set. Then, given an image with two generated captions from different models, subjects are asked to choose one caption that best describes the image. We received more than 9000 responses in total and the results are summarized in~\hyperref[fig:fig4]{Figure 4}. It can be seen that the majority of people consider the captions from G-GAN and especially our CAL better than those from the standard MLE method. This illustrates that despite both adversarial models perform poorly on automatic metrics, the generated captions are of higher quality in terms of human views. In the comparison between CAL and G-GAN, our model can generates more human-like captions that receive more acknowledgements. This demonstrates that, by exploiting more comparative relevance information against ground truth and other captions instead of solely on image, the proposed CAL effectively improves the caption generator and achieves better captions. We include failure analysis in \hyperref[sec:appendix3]{Appendices A.3}. 

\begin{figure*}[t]
 \centering
  \begin{tabular}{l  l  l  l  l }
    \hline
    \noalign{\smallskip}
    \begin{minipage}{.03\textwidth}
    \begin{sideways}
    \scriptsize{Bathroom}
    \end{sideways}
    \end{minipage}
    &
    \begin{minipage}{.21\textwidth}
      \includegraphics[width=\linewidth, height=0.60\linewidth]{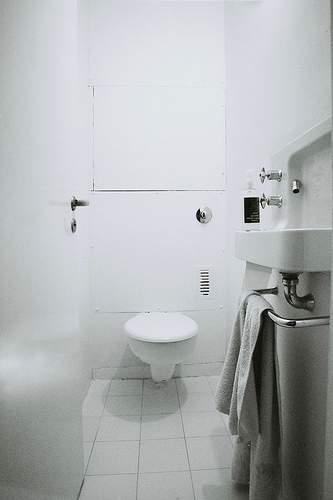}
    \end{minipage}
    &
  \begin{minipage}{.21\textwidth}
      \includegraphics[width=\linewidth, height=0.60\linewidth]{./example_figures/example5-12.jpg}
    \end{minipage}
    &
    \begin{minipage}{.21\textwidth}
      \includegraphics[width=\linewidth, height=0.60\linewidth]{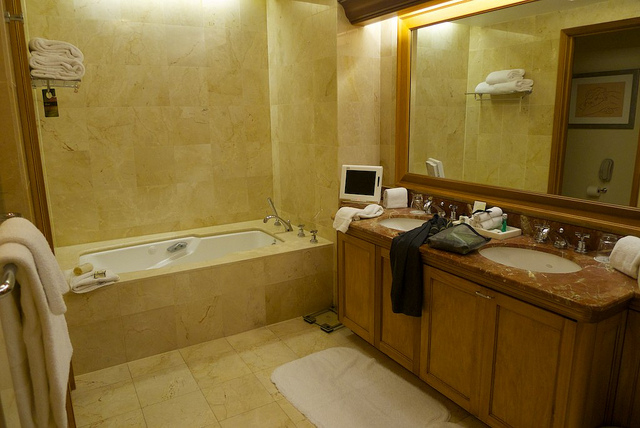}
    \end{minipage}
    &
    \begin{minipage}{.21\textwidth}
      \includegraphics[width=\linewidth, height=0.60\linewidth]{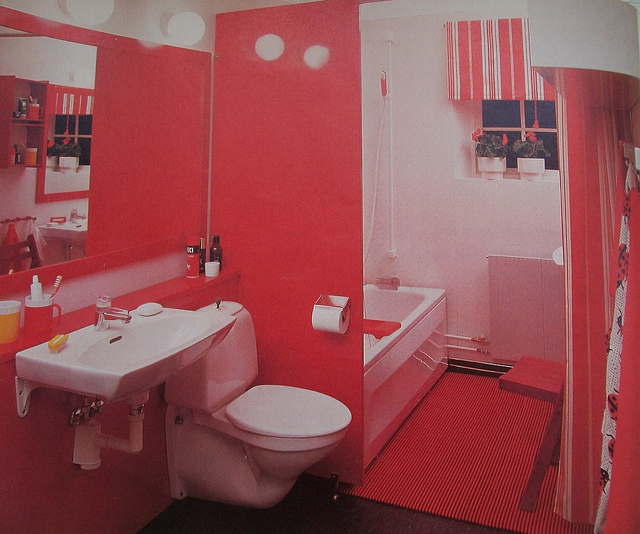}
    \end{minipage}

    \\
    \noalign{\smallskip}   
    \begin{minipage}{.03\textwidth}
    \begin{sideways}
    \scriptsize{MLE}
    \end{sideways}
    \end{minipage}
    &
    \begin{minipage}{.19\textwidth} \scriptsize
      a bathroom with a toilet and a sink
    \end{minipage}
    &
    \begin{minipage}{.19\textwidth} \scriptsize
      a bathroom with a toilet and a sink
    \end{minipage}
    &
    \begin{minipage}{.19\textwidth} \scriptsize
      a bathroom with a sink and a mirror
    \end{minipage}
    &
    \begin{minipage}{.19\textwidth} \scriptsize
      a bathroom with a sink and a toilet
    \end{minipage}
    
    \\ \noalign{\smallskip}  \hdashline
    \noalign{\smallskip}    
    \begin{minipage}{.03\textwidth}
    \begin{sideways}
    \scriptsize{G-GAN}
    \end{sideways}
    \end{minipage}
    &
    \begin{minipage}{.19\textwidth} \scriptsize
      a bathroom with a white toilet and tiled walls
    \end{minipage}
    &
    \begin{minipage}{.19\textwidth} \scriptsize
      a restroom with a toilet sink and shower 
    \end{minipage}
    &
    \begin{minipage}{.19\textwidth} \scriptsize
      a bathroom with a white bathtub and two sinks and a mirror 
    \end{minipage}
    &
    \begin{minipage}{.19\textwidth} \scriptsize
      a pink restroom with a toilet inside of it
    \end{minipage}
    
    \\ \noalign{\smallskip}  \hdashline
    \noalign{\smallskip}   
    \begin{minipage}{.03\textwidth}
    \begin{sideways}
    \scriptsize{CAL}
    \end{sideways}
    \end{minipage}
    &
    \begin{minipage}{.19\textwidth} \scriptsize
      a toilet sits inside of a bathroom next to a wall
    \end{minipage}
    &
    \begin{minipage}{.19\textwidth} \scriptsize
      a narrow bathroom with a toilet sink and a shower with dirty walls 
    \end{minipage}
    &
    \begin{minipage}{.19\textwidth} \scriptsize
      a clean bathroom with a large sink bathtub and a mirror 
    \end{minipage}
    &
    \begin{minipage}{.19\textwidth} \scriptsize
      a pink bathroom with a sink toilet and mirror
    \end{minipage}
    \\
    \noalign{\smallskip}
    \hline
    
    \noalign{\smallskip}
    
    \begin{minipage}{.03\textwidth}
    \begin{sideways}
    \scriptsize{Pizza}
    \end{sideways}
    \end{minipage}
    
    &
    \begin{minipage}{.21\textwidth}
      \includegraphics[width=\linewidth, height=0.60\linewidth]{./example_figures/example5-21.jpg}
    \end{minipage}
    &
  \begin{minipage}{.21\textwidth}
      \includegraphics[width=\linewidth, height=0.60\linewidth]{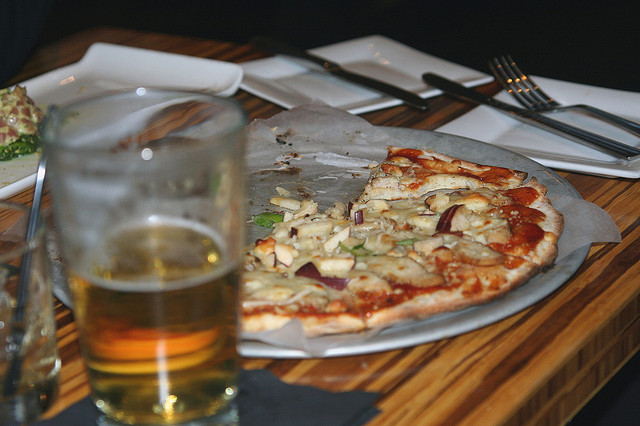}
    \end{minipage}
    &
    \begin{minipage}{.21\textwidth}
      \includegraphics[width=\linewidth, height=0.60\linewidth]{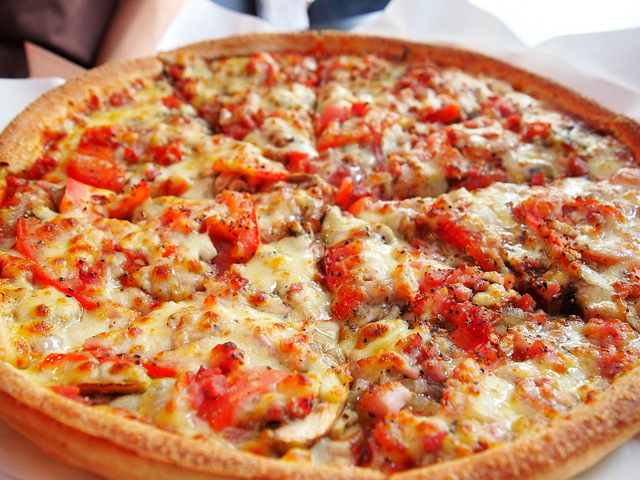}
    \end{minipage}
    &
    \begin{minipage}{.21\textwidth}
      \includegraphics[width=\linewidth, height=0.60\linewidth]{./example_figures/example5-24.jpg}
    \end{minipage}
    
    \\
    \noalign{\smallskip}   
    \begin{minipage}{.03\textwidth}
    \begin{sideways}
    \scriptsize{MLE}
    \end{sideways}
    \end{minipage}
    &
    \begin{minipage}{.19\textwidth} \scriptsize
      a pizza sitting on top of a white plate
    \end{minipage}
    &
    \begin{minipage}{.19\textwidth} \scriptsize
      a pizza sitting on top of a white plate
    \end{minipage}
    &
    \begin{minipage}{.19\textwidth} \scriptsize
      a close up of a pizza on a table
    \end{minipage}
    &
    \begin{minipage}{.19\textwidth} \scriptsize
      a pizza sitting on top of a pan
    \end{minipage}
    
    \\ \noalign{\smallskip}  \hdashline
    \noalign{\smallskip}    
    \begin{minipage}{.03\textwidth}
    \begin{sideways}
    \scriptsize{G-GAN}
    \end{sideways}
    \end{minipage}
    &
    \begin{minipage}{.19\textwidth} \scriptsize
      a pizza on a plate on a wooden table
    \end{minipage}
    &
    \begin{minipage}{.19\textwidth} \scriptsize
      a pizza sitting on a plate next to a glass of wine
    \end{minipage}
    &
    \begin{minipage}{.19\textwidth} \scriptsize
      the pizza is covered with cheese and tomatoes
    \end{minipage}
    &
    \begin{minipage}{.19\textwidth} \scriptsize
      a close up of a sliced pizza on a plate
    \end{minipage}
    
    \\ \noalign{\smallskip}  \hdashline
    \noalign{\smallskip}    
    \begin{minipage}{.03\textwidth}
    \begin{sideways}
    \scriptsize{CAL}
    \end{sideways}
    \end{minipage}
    &
    \begin{minipage}{.19\textwidth} \scriptsize
      a cheese pizza on a plate sits on a table
    \end{minipage}
    &
    \begin{minipage}{.19\textwidth} \scriptsize
      a plate of pizza and a glass of beer on the table 
    \end{minipage}
    &
    \begin{minipage}{.19\textwidth} \scriptsize
      a pizza topped with lots of toppings is ready to be cut 
    \end{minipage}
    &
    \begin{minipage}{.19\textwidth} \scriptsize
     a partially eaten pizza is being cooked on a pan
    \end{minipage}
    \\
    \noalign{\smallskip}
    \hline
    
    \noalign{\smallskip}
    
    \begin{minipage}{.03\textwidth}
    \begin{sideways}
    \scriptsize{Car}
    \end{sideways}
    \end{minipage}
    
    &
    \begin{minipage}{.21\textwidth}
      \includegraphics[width=\linewidth, height=0.60\linewidth]{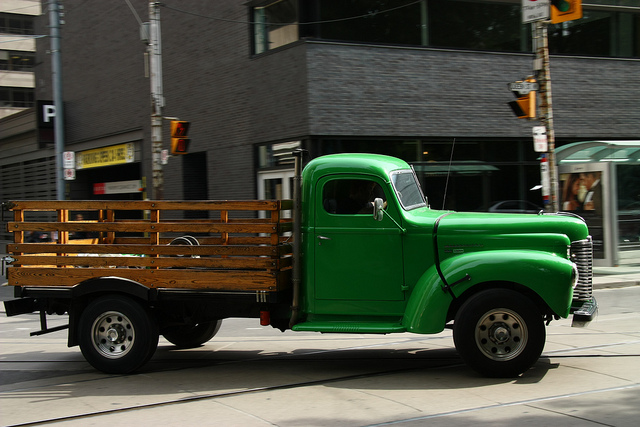}
    \end{minipage}
    &
  \begin{minipage}{.21\textwidth}
      \includegraphics[width=\linewidth, height=0.60\linewidth]{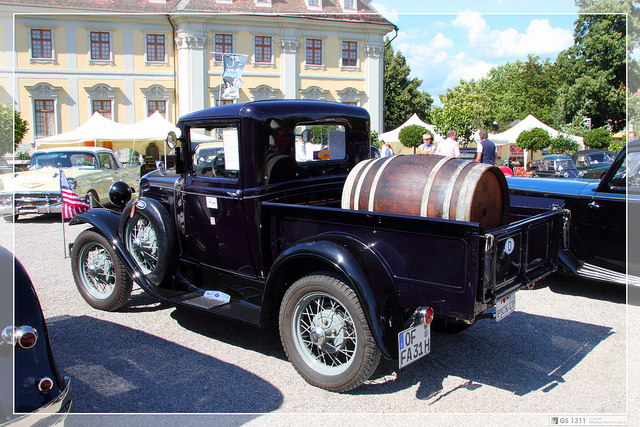}
    \end{minipage}
    &
    \begin{minipage}{.21\textwidth}
      \includegraphics[width=\linewidth, height=0.60\linewidth]{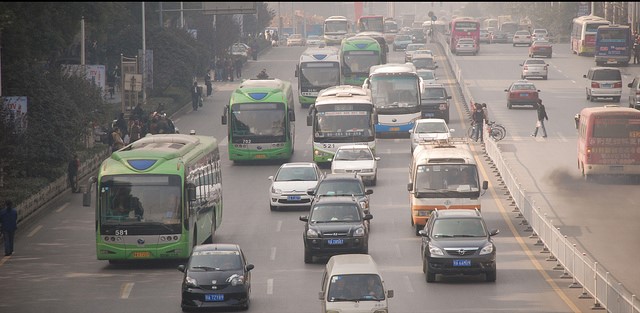}
    \end{minipage}
    &
    \begin{minipage}{.21\textwidth}
      \includegraphics[width=\linewidth, height=0.60\linewidth]{./example_figures/example5-34.jpg}
    \end{minipage}
    
    \\
    \noalign{\smallskip}   
    \begin{minipage}{.03\textwidth}
    \begin{sideways}
    \scriptsize{MLE}
    \end{sideways}
    \end{minipage}
    &
    \begin{minipage}{.19\textwidth} \scriptsize
      a green truck parked in a parking lot
    \end{minipage}
    &
    \begin{minipage}{.19\textwidth} \scriptsize
      a black truck is parked in a parking lot
    \end{minipage}
    &
    \begin{minipage}{.19\textwidth} \scriptsize
      a group of buses driving down a street
    \end{minipage}
    &
    \begin{minipage}{.19\textwidth} \scriptsize
      a city bus stopped at a bus stop
    \end{minipage}
    
    \\ \noalign{\smallskip}  \hdashline
    \noalign{\smallskip}    
    \begin{minipage}{.03\textwidth}
    \begin{sideways}
    \scriptsize{G-GAN}
    \end{sideways}
    \end{minipage}
    &
    \begin{minipage}{.19\textwidth} \scriptsize
      a green garbage truck in a business district
    \end{minipage}
    &
    \begin{minipage}{.19\textwidth} \scriptsize
      an antique black car sitting in a parking lot
    \end{minipage}
    &
    \begin{minipage}{.19\textwidth} \scriptsize
      a city street filled with taxis and buses
    \end{minipage}
    &
    \begin{minipage}{.19\textwidth} \scriptsize
      people are waiting in line as the bus travel down the road
    \end{minipage}
    
    \\ \noalign{\smallskip}  \hdashline
    \noalign{\smallskip}   
    \begin{minipage}{.03\textwidth}
    \begin{sideways}
    \scriptsize{CAL}
    \end{sideways}
    \end{minipage}
    &
    \begin{minipage}{.19\textwidth} \scriptsize
      a large green truck driving past a tall building
    \end{minipage}
    &
    \begin{minipage}{.19\textwidth} \scriptsize
      an old style truck parked in a parking space near a building 
    \end{minipage}
    &
    \begin{minipage}{.19\textwidth} \scriptsize
      the city buses are driving through the traffic
    \end{minipage}
    &
    \begin{minipage}{.19\textwidth} \scriptsize
     people gather to a street where a bus get ready to board
    \end{minipage}
    \\
    \noalign{\smallskip}
    \hline
  \end{tabular}
  \caption{Qualitative results illustrate that adversarial models, especially our proposed CAL, can generate more diverse descriptions.}
  \label{fig:fig5}
\end{figure*}

\begin{table}[t]
\setlength{\tabcolsep}{11pt}
\small
\centering
	\begin{tabular}{c c c c|c}
	\hline
	\multicolumn{4}{c|}{Adversarial Model} &  \multirow{2}{*}{Diversity} \\
	\hhline{----~}
	bs & samp & noise & compa & \\ \hline
	\checkmark & & & & 7.078 \\ \hline
	 & \checkmark & & & 7.331 \\ \hline
	 & \checkmark & \checkmark & & 7.784 \\ \hline
	 & \checkmark & \checkmark & \checkmark & 8.845 \\ \hline
	\end{tabular}
\caption{Ablation study of caption diversity of our adversarial model. \textit{bs} and \textit{samp} indicate beam search and sampling decoding respectively. \textit{noise} denotes adding noise vectors in decoding. \textit{compa} represents our discriminator with comparative learning.}
\label{tab:ablation_diversity}
\end{table}

\vspace{2mm}
\noindent\textbf{Diversity.}\hspace{1mm} To compare the capabilities of generating diverse expressions, we measure the variances of captions from different models across images. All the embedding features are extracted using the same text-encoder in our framework. Besides estimating the variance across all the images in the test set, we also inspect the variances inside different image categories. Particularly, We use the K-means method to cluster the input image features, and select six clusters with high-frequency topics. All the results are summarized in~\hyperref[table:tab2]{Table 2}. It can be seen that despite the MLE method performs well on automatic metrics, the variance of captions is relatively lower across different images. As shown in~\hyperref[fig:fig5]{Figure 5}, the MLE model often generates similar expressions and meanings within one category, even if the images are distinct.
\begin{figure*}[t]
 \centering
  \begin{tabular}{l  l  l  l  l }
    \hline
    \noalign{\smallskip}
    \begin{minipage}{.03\textwidth}
    \begin{sideways}
    \scriptsize{}
    \end{sideways}
    \end{minipage}
    &
    \begin{minipage}{.21\textwidth}
      \includegraphics[width=\linewidth, height=0.60\linewidth]{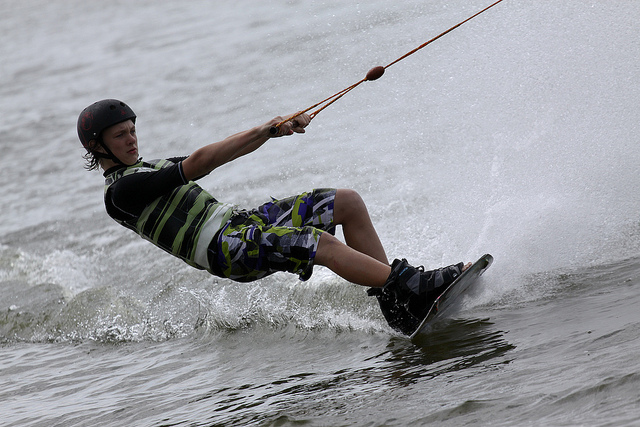}
    \end{minipage}
    &
  \begin{minipage}{.21\textwidth}
      \includegraphics[width=\linewidth, height=0.60\linewidth]{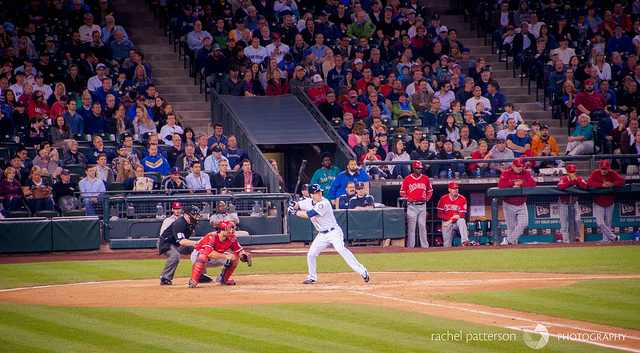}
    \end{minipage}
    &
    \begin{minipage}{.21\textwidth}
      \includegraphics[width=\linewidth, height=0.60\linewidth]{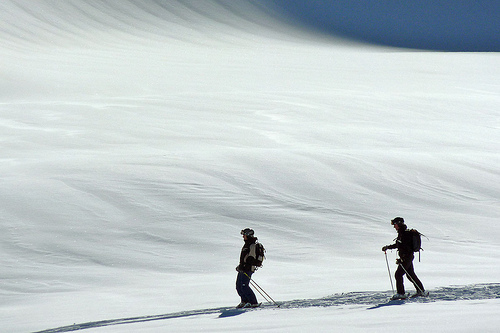}
    \end{minipage}
    &
    \begin{minipage}{.21\textwidth}
      \includegraphics[width=\linewidth, height=0.60\linewidth]{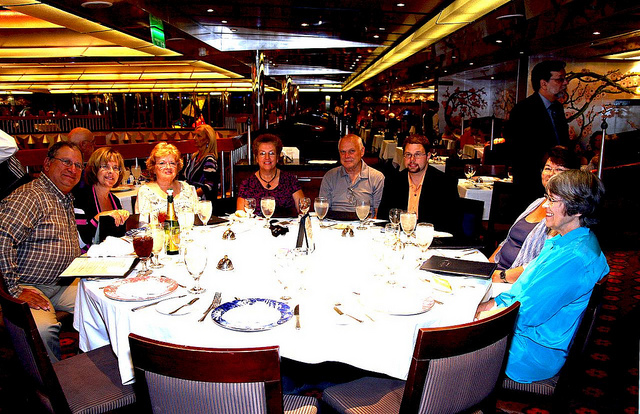}
    \end{minipage}

    \\
    \noalign{\smallskip}   
    \begin{minipage}{.03\textwidth}
    \scriptsize{$z_1$}
    \end{minipage}
    &
    \begin{minipage}{.19\textwidth} \scriptsize
      a man holding a rope while skateboarding a small wave
    \end{minipage}
    &
    \begin{minipage}{.19\textwidth} \scriptsize
      a baseball game is in progress with the large crowd watching
    \end{minipage}
    &
    \begin{minipage}{.19\textwidth} \scriptsize
      twin skiers stand on skis on a snowy hillside
    \end{minipage}
    &
    \begin{minipage}{.19\textwidth} \scriptsize
      a crowd of people are having a meal and drinking beer at a table
    \end{minipage}
    
    \\ \noalign{\smallskip}  \hdashline
    \noalign{\smallskip}    
    \begin{minipage}{.03\textwidth}
    \scriptsize{$z_2$}
    \end{minipage}
    &
    \begin{minipage}{.19\textwidth} \scriptsize
      a person is trying to flip while skiing water
    \end{minipage}
    &
    \begin{minipage}{.19\textwidth} \scriptsize
      a person in jersey holding a baseball bat in swing position 
    \end{minipage}
    &
    \begin{minipage}{.19\textwidth} \scriptsize
      two people in ski gear standing on a snowy
slope 
    \end{minipage}
    &
    \begin{minipage}{.19\textwidth} \scriptsize
      a group of people are eating outside at a restaurant
    \end{minipage}
    
    \\ \noalign{\smallskip}  \hdashline
    \noalign{\smallskip}   
    \begin{minipage}{.03\textwidth}
    \scriptsize{$z_3$}
    \end{minipage}
    &
    \begin{minipage}{.19\textwidth} \scriptsize
      there is a male surfer that is riding a wave
    \end{minipage}
    &
    \begin{minipage}{.19\textwidth} \scriptsize
      a batter catcher and umpire during a baseball game 
    \end{minipage}
    &
    \begin{minipage}{.19\textwidth} \scriptsize
      two skiers on a thick snow covered mountain 
    \end{minipage}
    &
    \begin{minipage}{.19\textwidth} \scriptsize
      many people are sitting on tables at a restaurant
    \end{minipage}
    \\
    \noalign{\smallskip}
    \hline
  \end{tabular}
  \caption{Qualitative results illustrate the images captioned by CAL with different random vector $z$.}
  \label{fig:fig6}
\end{figure*}

In contrast to the MLE model, both adversarial models, especially our proposed CAL, can generate more diverse captions with respects to distinct images. 
G-GAN uses a binary discriminator to separates the decisions on machine-generated and human-written captions. Compared to G-GAN, our network trained by comparative learning binds the information of human-written captions which possess highest diversity characteristics as indicated in~\hyperref[table:tab2]{Table 2}. The comparative learning also encourages the distinctiveness of the generated captions by suppressing the cr-scores of mismatched captions, especially for those from akin images. These allow our caption generator to produce more descriptive captions for different images. As expected, the variance of captions from our model is larger than that from G-GAN across all the images. Similar trends can be observed inside different categories. We also notice that the word usage in our CAL is more diverse and akin to that of human than other baselines (\hyperref[sec:appendix2]{Appendices A.2}). These suggest that our proposed CAL has better generative capability than the baseline G-GAN and helps bridge the gap between machine-generated and human-written captions. \hyperref[fig:fig6]{Figure 6} shows that the CAL model is able to generate diverse captions for each image.

\vspace{2mm}
\noindent\textbf{Ablation.} We study the diversity effect of each component in our network on MSCOCO val set. The results are summarized in~\hyperref[tab:ablation_diversity]{Table 3}, where we can see that the sampling decoding and noise vectors bring a certain amount of diversity. The proposed comparative relevance discriminator compares different captions and maximizes the scores of the generated captions among a set of references, resulting in an even larger diversity gain.

\vspace{2mm}
\noindent\textbf{Network Effectiveness.} We further investigate the effectiveness of adversarial models by a caption-image matching experiment~\cite{Dai2017Towards,dai2017contrastive}. Specifically, if all the generated captions have enough diversity and the adversarial discriminator is good enough to distinguish related and unrelated image-caption pairs, the corresponding image could be easily retrieved by the discriminator when given its own caption. For each adversarial model, we can use its generated caption as a query to rank all images, based on the similarity scores from corresponding discriminators. Then a recall ratio can be calculated by inspecting the top-k resulting images in the ranked list. Since the MLE model is not an adversarial model, we use the discriminator from G-GAN to retrieve the generated captions, providing a baseline for comparison. 

\begin{table}[t]
\small
\begin{center}
\setlength{\tabcolsep}{7.5pt}
\begin{tabular}{ccccc}
\hline\noalign{\smallskip}
Model & R@1 & R@3 & R@5 & R@10  \\
\hline\noalign{\smallskip}
MLE & 3.03 & 8.67 & 12.75 & 20.54  \\
\hline\noalign{\smallskip}
G-GAN w/o reg. & 14.24 & 30.28 & 40.11 & 56.13  \\
\hline\noalign{\smallskip}
G-GAN & 16.07 & 33.66 & 43.80 & 59.74  \\
\hline\noalign{\smallskip}
CAL (ours) & \textbf{18.81} & \textbf{36.56} & \textbf{46.84} & \textbf{62.57}  \\
\noalign{\smallskip}
\hline
\end{tabular}
\caption{ Caption-image matching comparison evaluated on MSCOCO test set. Captions are all self-generated by each model. \textit{G-GAN w/o reg.} denotes the G-GAN model without the regularization term in the discriminator. The recall ratio is calculated by ranking discriminator's scores based on caption-image pairs. }
\label{table:tab4}
\end{center}
\end{table}

\hyperref[table:tab4]{Table 4} shows the performance comparison of different models. Although captions from MLE commonly describe images well, they are less diverse for different images, resulting in a poor retrieval performance. Meanwhile, our proposed CAL outperforms all the other models, including the adversarial model G-GAN. This further demonstrates that CAL can produce more diverse captions for all images. In \hyperref[table:tab4]{Table 4}, it is noteworthy that the G-GAN model needs an regularization term to sustain better semantic relevance of captions. Without such regularization, our CAL model still improves discernibility on caption-image pairs. It proves that the cr-discriminator in our proposed network can provide more accurate rewards during adversarial training, leading to a better caption generator. 
\section{Conclusions}
We presented a comparative adversarial learning network for generating diverse captions across images. A novel comparative learning schema is proposed for the discriminator, which better assesses the quality of captions by comparing with other captions. Thus more caption properties including correctness, naturalness, and diversity can be taken into consideration. This in turn benefits the caption generator to effectively exploit inherent characteristics inside human languages and generate more diverse captions. We also proposed a new caption diversity metric in the semantic level across images. Experimental results clearly demonstrate that our proposed method generates better captions in terms of both accuracy and diversity across images.

\bibliographystyle{aaai}
\bibliography{egbib}

\newpage\null\thispagestyle{empty}\newpage
\appendix
\section{Appendices}
\vspace{4mm}
\subsection{Implementation Details.}
\label{sec:appendix1}

Following~\cite{karpathy2015deep}, we convert all the captions to lowercases and remove its non-alphabet characters. We also discard the tokens with frequency less than 5 in the training dataset, resulting in a vocabulary size of 8,791. Both image encoders $F_{G_\theta}$ and $F_{D_\phi}$ in the generator and discriminator are implemented using \textit{ResNet}~\cite{he2016deep} with 152 layers, separately. The image activations in the \textit{pool5} layer are extracted, yielding 2048-dimensional image features. Noise vector $z$ with 100-dimensions is sampled from a uniform distribution. All the image features are projected to 512 dimensions by fully connected layers. The text-decoder in the generator and the text-encoder in the discriminator are all implemented using LSTMs with 512 hidden nodes. We use the last hidden activations from the text-encoder as text feature, which shares the same dimension with the projected image feature. 
\subsection{Diversity visualization}
\label{sec:appendix2}

\begin{figure*}[htbp]
\centering
\includegraphics[height=6.0cm]{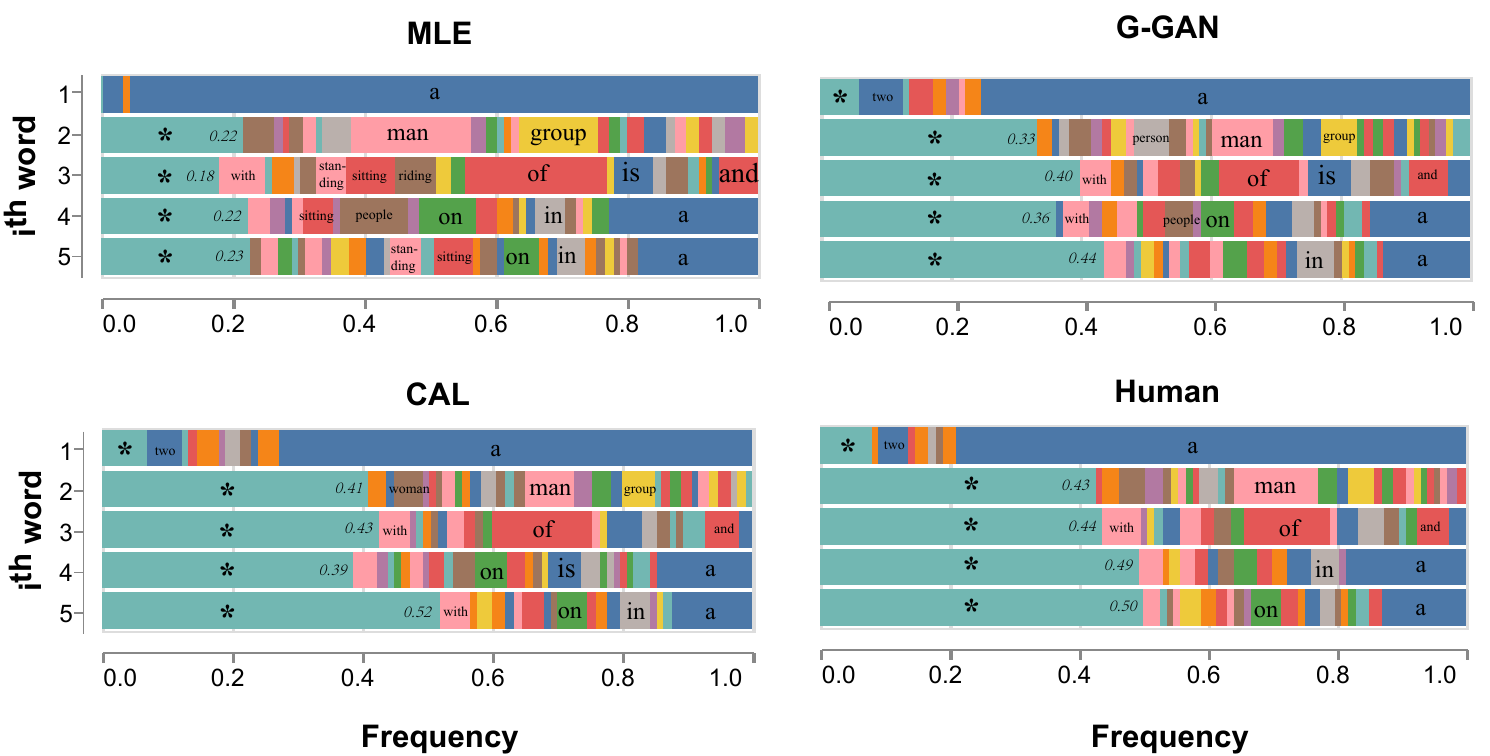}
\caption{Visualization of word diversity produced by different models. For each subgraph, the $i^{th}$ row represents the $i^{th}$ word's frequency distribution in all the generated captions. Different colors denote different words, and the width of each region is proportional to the frequency of the corresponding word. We only plot the first five words for easy readability. We mark the words of high frequency in the figure. The words of low frequency ($ < 0.5\%$) are merged into \textit{the others} (denoted as \textbf{*}) category. Larger proportion of \textbf{*} means more chances of using diverse words (or long tail words) in the vocabulary. The decimal in each \textbf{*} region denotes its proportion value for easy comparison. }
\label{fig:fig7}
\end{figure*}

As distinctive image contents are described by specific words, caption diversity across images could be visualized by diverse word usages. For this purpose, we can inspect the word usage frequency at each position $t$ of the generated captions:
\begin{align*}
p(w_t) & = \mathbb{E}_{w_{t-1}} p(w_t | w_{t-1}) p(w_{t-1}), \hspace{2mm} t\in(1, T)\\
\end{align*}
where $w_0$ only takes one word which is the "START" token and thus $p(w_0="\text{START}")=1$, and the image notation $I$ is ignored for simplification. The above probability can be estimated by the Markov Chain method over the generated vocabulary distribution (Equation 2). However, as the word space is too large, it is practically difficult to calculate the frequency distribution for all the words. Instead, we use a sampling method to approximate the frequency of the observed words in the generated captions: 
\begin{equation*}
p(w_t) \simeq count(w_t) / count(\bar{w}_{t})\hspace{2mm} t\in(1, T)
\end{equation*}
where $count(\bar{w}_{t})$ denotes the total count of all the observed words at position $t$. Ideally, diverse word usage implies that each word in the vocabulary is used less repeatedly in caption generation across different images, leading to lower $p(w_t)$ for each word. 

To visualize the word usage frequency, we sampled 300 images from the test set and visualize the statistics in \hyperref[fig:fig7]{Figure 7}. Here we chose 300 but not more images for the sake of visualization clarity.
As can be seen, the MLE-generated captions usually pick up fewer content words such as "sitting", "riding", and "standing", regardless of different and distinctive image contents. Meanwhile, its corresponding \textbf{*} regions are much narrower than those in other models, meaning that it rarely uses other words in the vocabulary. In contrast, although our CAL model also uses the same function words (e.g. "a", "the", "of", "is", etc.), most of the used content words are less identical and contribute to much wider \textbf{*} regions. We also find that our CAL uses more adjectives and adverbs in generated captions.

By comparison, we can conclude that the word frequency distribution of our CAL is more akin to that of Human. This demonstrates that our CAL model has more diverse word usages than the baseline G-GAN, resulting in more distinctive captions across images.

\begin{figure*}[t]
 \centering
  \begin{tabular}{l l  l }
    \hline
    \noalign{\smallskip}
    \begin{minipage}{.05\textwidth}
    \end{minipage}
    &
    \begin{minipage}{.42\textwidth}
      \includegraphics[width=0.87\linewidth]{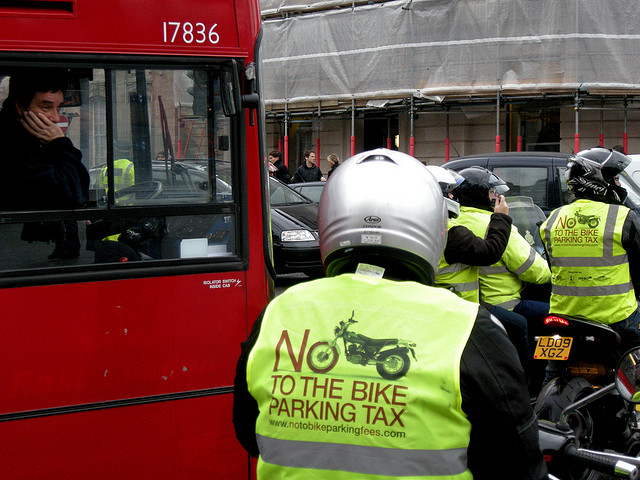}
    \end{minipage}
    &
  	\begin{minipage}{.42\textwidth}
      \includegraphics[width=\linewidth]{./example_figures/fail-example2.jpg}
    \end{minipage}
    \\
    \noalign{\smallskip}
    \hline
    
    \noalign{\smallskip}
    \begin{minipage}{.05\textwidth}

    \end{minipage}
    & 
    \begin{minipage}{.35\textwidth}
      a police officer directing traffic at a bus stop
    \end{minipage}
    & 
  	\begin{minipage}{.40\textwidth}
      a display store of various old speckled benches
    \end{minipage}
    \\
    \noalign{\smallskip}
    \hline

  \end{tabular}
  \caption{Failure examples from the proposed network.}
  \label{fig:fig8}
\end{figure*}

\subsection{Failure analysis}
\label{sec:appendix3}
For some images with complex content, we find the CAL-generated captions are imprecise or defective in describing their contents. One possible reason is that if a complicate image does not have a focused topic, its ground truths are normally divergent for different aspects of the image. As a result, during comparative adversarial learning, the caption generator can not simultaneously capture all the opposed details from ground truths. Additionally, to generate more descriptive and diverse captions for images, our framework bears risks that involving some incorrect details. \hyperref[fig:fig8]{Figure 8} shows some failure examples from our method. We will consider these problems in our future study.

\end{document}